\documentclass[sigconf,arxiv,nonacm]{acmart}

\AtBeginDocument{%
  \providecommand\BibTeX{{%
    \normalfont B\kern-0.5em{\scshape i\kern-0.25em b}\kern-0.8em\TeX}}}

\setcopyright{acmcopyright}
\copyrightyear{2021}
\acmYear{2021}
\acmDOI{11.1111/1111111.1111111}

\acmConference[Jerusalem '21]{The Fourteenth ACM International Conference on Web Search and Data Mining (WSDM ’21)}{March 8–12, 2021}{Jerusalem, Israel}
\acmBooktitle{Jerusalem '21: The Fourteenth ACM International Conference on Web Search and Data Mining (WSDM ’21), March 8–12, 2021, Jerusalem, Israel}
\acmPrice{15.00}
\acmISBN{978-1-4503-XXXX-X/18/06}

\usepackage{subfigure}
\begin{document}

\title{HeteGCN: Heterogeneous Graph Convolutional Networks for Text Classification}

\author{Rahul Ragesh, Sundararajan Sellamanickam, Arun Iyer*, Ram Bairi*, Vijay Lingam}

\authornote{equal contribution}
\email{t-rarage,ssrajan,ariy,ram.bairi,t-vili@microsoft.com}

\affiliation{%
  \institution{Microsoft Research India}
  \city{Bengaluru, India}
}
\renewcommand{\shortauthors}{Rahul Ragesh, et al.}
\newcommand{\bfx}{\mathbf{X}}
\newcommand{\bftx}{\mathbf{TX}}
\newcommand{\bff}{\mathbf{F}}
\newcommand{\bfn}{\mathbf{N}}
\newcommand{\bfl}{\mathbf{L}}
\newcommand{\bfa}{\mathbf{A}}
\newcommand{\bfe}{\mathbf{E}}
\newcommand{\bfw}{\mathbf{W}}
\newcommand{\bfu}{\mathbf{U}}
\newcommand{\bfsu}{\mathbf{u}}
\newcommand{\bfv}{\mathbf{V}}
\newcommand{\bfsv}{\mathbf{v}}
\newcommand{\bfi}{\mathbf{I}}
\newcommand{\bfp}{\mathbf{P}}
\newcommand{\tgcn}{\textsc{TextGCN}\:}
\newcommand{\gcn}{\textsc{GCN}\:}
\newcommand{\hgcn}{\textsc{HeteGCN}\:}
\newcommand{\pte}{\textsc{PTE}\:}
\newcommand{\lgcn}{\textsc{LightGCN}\:}
\newcommand{\lr}{\textsc{LR}\:}

\newcommand{\bow}{\textsc{BoW}\:}
\newcommand{\tfidf}{\textsc{TF-IDF}\:}

\begin{abstract}
 We consider the problem of learning efficient and inductive graph convolutional networks for text classification with a large number of examples and features. Existing state-of-the-art graph embedding based methods such as predictive text embedding (\pte) and \tgcn have shortcomings in terms of predictive performance, scalability and inductive capability. To address these limitations, we propose a heterogeneous graph convolutional network (\hgcn) modeling approach that unites the best aspects of \pte and {\tgcn}~together. The main idea is to learn feature embeddings and derive document embeddings using a \hgcn architecture with different graphs used across layers. We simplify \tgcn by dissecting into several \hgcn models which (a) helps to study the usefulness of individual models and (b) offers flexibility in fusing learned embeddings from different models. In effect, the number of model parameters is reduced significantly, enabling faster training and improving performance in small labeled training set scenario. Our detailed experimental studies demonstrate the efficacy of the proposed approach.              
\end{abstract}


\maketitle

\section{Introduction}
\label{sec:introduction}
Text classification has been an important class of machine learning problems for several decades with challenges arising from different dimensions including a large number of documents, features and labels, label sparsity, availability of unlabeled data and side information, training and inference speed, type of classification problem (binary, multi-class, multi-label). Many statistical models~\cite{naivebayes, svm, svmbayes, reviewcharu} and machine learning techniques~\cite{cddual, sslkeerthi, multiclasssvm, dlxmltc} have been proposed to address these challenges. 

Traditional document classification approaches use Bag-of-Words (\bow) sparse representation of documents~\cite{naivebayes} and contributed towards designing models for binary, multi-class and multi-label classification problems~\cite{svm, svmbayes, sslkeerthi, dlxmltc} and speeding-up learning algorithms~\cite{cddual} for large scale applications. Semi-supervised learning methods~\cite{svm, cotraining} became an important area of research with the availability of an extensive collection of unlabeled data (e.g., web pages). Furthermore, web pages and publications brought in rich information through \textit{link graphs} enabling the use of auxiliary information. Development of new graph-based learning methodologies~\cite{linkbased, coregularization} started with the goal of improving classifier model performance. 

For nearly a decade, there has been a surge in the development of learning representation of text data using distributed representation models~\cite{word2vec, glove, para2vec} and deep learning models (e.g., convolutional and recurrent neural networks). See~\cite{minaeereview} for a comprehensive review. These models achieve superior performance compared to traditional models and exploit large volume of available unlabeled data to learn word, sentence and document embeddings. The key idea is to learn better representation (\textit{embedding) } for documents that help to get significantly improved performance using even off-the-shelf linear classifier models. Recently, there has been tremendous progress in learning embeddings for entities in relational graphs~\cite{deepwalk, metapath2vec, line, pte} and designing graph convolutional and neural network models~\cite{gcn} that exploit rich relational information present among entities in the graph. Many variants of GCNs and GNNs~\cite{gcn, textgcn, lightgcn, hine, kgat} have been proposed and explored for solving classification and recommendation problems. 

Our interest lies in learning text classifier models that make use of underlying relations among words and documents. In Predictive Text Embedding~\cite{pte} modeling approach, a document corpus is viewed as a heterogeneous graph that encodes relationship among words, documents and labels. The method learns embeddings for words using unsupervised learning techniques using a large volume of unlabeled data and some labeled data. It derives document embedding using learned word embeddings and learns a simple linear classifier model for class prediction. Like \pte, \tgcn~\cite{textgcn} uses a heterogeneous graph but learns a text classifier model with a graph convolutional network and it outperforms many popular neural network models~\cite{fasttext, rnntext} and \pte on several benchmark datasets. \pte's modelling approach is simple, efficient and inductive, but it suffers in performance compared to the complex, slow and transductive \tgcn. Therefore, we focus on designing a text classification modeling approach that unites the best aspects of \pte and \tgcn together. 

A key contribution of this work is a proposal to compose different heterogeneous graph convolution networks (\hgcn) with individual graphs used in \tgcn. Unlike traditional \gcn and \tgcn, \hgcn makes use of \textit{compatible} graphs across layers. They are simple and efficient as the number of model parameters is reduced significantly, enabling faster training and better generalization performance when the number of labeled training examples are small. Our \hgcn modeling approach helps to the understand effectiveness of different \hgcn model variants and their usefulness in different application scenarios (e.g., availability of auxiliary information, compute and storage constrained settings). \hgcn offers flexibility in designing networks with fusion and shared learning capabilities.

Following \pte, we suggest a simple idea of using learned feature embeddings from both \tgcn and \hgcn for inductive inference. Our work also raises a few research questions from inductive inference perspective. Further, \lgcn work talks about simplifying \gcn for recommendation application~\cite{lightgcn}. We show how \lgcn can be used for the text classification task as a competitive baseline for \tgcn. 

We conduct a detailed experimental study on several benchmark datasets and compare \hgcn with many state-of-the-art methods. \hgcn outperforms these methods when the number of labeled training examples is small and gives as 
high as $2-8\%$ improvement over \tgcn and \pte for several datasets. It provides competitive performance in transductive large labeled training data scenario. In inductive setting, we find that the idea of using only learned feature embeddings is quite effective and \hgcn achieves significantly improved performance with $3-7\%$ lifts on several datasets compared to \tgcn and \pte. Training time comparison shows that \hgcn is faster than \tgcn by $2-9$ times. Overall, we demonstrate how \hgcn unites the best aspects of \pte and \tgcn by offering a high-performance text classification solution with lower model complexity, faster training and inductive capabilities. 


The paper is organized as follows. We present notation, problem formulation and background for graph embedding methods in Sections~\ref{sec:preliminaries} \& \ref{sec:background}, our approach in Section~\ref{sec:proposedapproach}, experimental details and results in Section~\ref{sec:experiments} followed by related work in Section~\ref{sec:relatedwork}. The paper ends with discussion on investigations and possible extensions in Section~\ref{sec:discussion} followed by conclusion.    

\section{Notation and Problem Formulation}
\label{sec:preliminaries}
We introduce notation used throughout the paper. We use $x_i \in {\mathcal R}^m$ to denote $m$-dimensional feature vector of $i^{th}$ example, $y_i \in \{1,\ldots,k\}$ to denote the corresponding binary representation of target class and $k$ to denote the number of classes. We use $\bfx$ to denote the \textit{document-feature} matrix of the entire corpus (where each row represents the feature vector of an example) and $\mathcal{L}_{tr}$ to denote the set of labeled training examples. We use $\bff$ to denote a \textit{feature-feature} relation matrix and $\bfn$ to denote a \textit{document-document} relation matrix. The relation matrices are either explicitly provided or implicitly constructed from $\bfx$. We view each matrix as a graph (and vice-versa) with rows and columns representing nodes. For example, $\bfx$ is a graph connecting document nodes and feature nodes with edge weights that may be set to \textsc{TF-IDF} scores. We use $\bfsv_i = Emb(d_i)$ ($i \in \{1,\ldots,n\}$) to denote the document embedding of the $i^{th}$ document and $\bfsu_j = Emb(f_j)$ to denote the embedding of the $j^{th}$ feature ($j \in \{1,\ldots,m\}$). The corresponding embedding matrices are denoted as $\bfv$ and $\bfu$ respectively. Finally, $\bfi_n$ and $\bfw$ denote the $n$ dimensional identity matrix and the network model weights respectively. We use $\bfp$ to denote the class probability distribution of documents over $k$ classes. The terms \textit{document} and \textit{example} are used interchangeably. Similarly, we use \textit{feature} and \textit{word} interchangeably. In some privacy sensitive applications, only hashed feature identifiers are available and pre-trained word embeddings (e.g.,\textsc{Word2Vec}) cannot be used. 

\textbf{Problem Formulation.} We are given a document corpus ($\bfx)$ with a set of labeled and unlabeled examples. Additionally, we may have access to $\bff$ and $\bfn$. Our goal is to learn feature embeddings ($\bfu$) using simple and efficient graph convolutional networks and learn an inductive classifier model that makes use of document embeddings ($\bfv$) computed using $\bfu$ to predict the target class. 



 \section{Background and Motivation}
 \label{sec:background}
 \textbf{Predictive Text Embedding (\pte).} Tang et al.~\cite{pte} proposed a semi-supervised embedding learning method that makes use of both labeled and unlabeled documents. They construct a heterogeneous text graph which is a composition of three graphs, document-word ($\bfx$), word-word ($\bff$) and word-label ($\bfl$). Note that $\bfx$ and $\bff$ are available for the entire corpus and $\bfl$ is constructed using only labeled training examples. The core idea is to learn \textit{shared} word embeddings ($\bfu$) by leveraging labeled information available through $\bfl$ jointly with $\bfx$ and $\bff$. 
 The embedding of any document is computed as an average of the embeddings of words present in the document. Finally, a linear (\textsc{Softmax}) classifier model is learned by minimizing cross-entropy loss function with labeled training examples: \begin{equation}
 {\mathcal M}(\bfw) = -\sum_{i \in {\mathcal L}_{tr}} \sum_{k} y_{i,k} \ln(p(y_{i,k}|\bfsv_i;\bfw))
 \label{eqn:training}
 \end{equation} Note that the embeddings and classifier model are learned \textit{sequentially}. \pte is efficient and has the inductive capability. 
 
 \textbf{Graph Convolutional Network .} Kipf et al.~\cite{gcn} proposed a semi-supervised learning method using graph convolutional networks. A graph convolutional network is a multi-layer network and each layer takes node embeddings ($\bfe^{(l)})$ of the $l^{th}$ layer as input and produces either embeddings ($\bfe^{(l+1)})$ for the $(l+1)^{th}$ layer or predictions for a given task (e.g., target class probability distribution) at the final layer. Formally, $\bfe^{(l+1)} = g(\bfa,\bfe^{(l)},\bfw^{(l)})$ where $\bfe^{(l)}$ and $\bfw^{(l)}$ denote the embedding and weight matrices of the $l^{th}$ layer respectively. $g(\cdot)$ is a transformation function (e.g., \textsc{ReLU} or \textsc{Softmax}), and $\bfe^{(0)}$ is the input feature matrix (e.g., $\bfx$) or 1-hot encoding representation of nodes (i.e., $\bfi_n$). $\bfa$ denotes an adjacency matrix and is fixed across layers. Like \pte, the model weights $\bfw$ are learned using only labeled training examples (see \ref{eqn:training}). However, \gcn is only transductive, as embeddings for unseen documents cannot be computed during inference. 
 
 \textbf{TextGCN}. Yao et al.\cite{textgcn} proposed a text graph convolutional network (\textsc{TextGCN}) method by using the matrix: 
 \begin{equation}
 \bfa = \begin{bmatrix}
 \bff & \bfx^T \\
 \bfx & {\bf 0}
 \end{bmatrix}
 \label{eqn:A}
\end{equation}
in \textsc{GCN} across layers. Note that it makes use of both \textit{word-word} and \textit{document-word} relational graphs excluding $\bfn$ (which may not be explicitly available). With a two-layer network, document-document relations are inferred and used along with word embeddings to obtain document embeddings. \textsc{TextGCN}~\cite{textgcn} achieved significantly improved accuracy performance over \pte. \tgcn does not make use of $\bfl$, and the gain mainly comes from using \gcn. Nevertheless, \tgcn has several limitations which we explain next. 

Consider the first layer output of \tgcn. With $\bfa$ as given in~\ref{eqn:A}, the feature and document embedding matrices are given by: 
{\small
\begin{align}
    \bfu^{(1)} = g(\bff \bfi_m \bfw_u^{(0)} + \bfx^T \bfi_n \bfw_v^{(0)}), \bfu^{(2)} = g((\bff \bfu^{(1)} + \bfx^T \bfv^{(1)}) \bfw_u^{(1)}) \label{eqn:wemb} \\ 
    \bfv^{(1)} = g(\bfx^T \bfi_n \bfw_v{(0)}), \bfp = \textsc{Softmax}(\bfx \bfu^{(1)} \bfw_v^{(1)}) \label{eqn:demb}
\end{align}
}
where 1-hot representation is used for features and documents. The subscripts in $\bfw_v^{(0)}$ and $\bfw_u^{(0)}$ denote embedding type in the first (\textit{aka} input) layer. In other layers, we use the subscript in the weight matrix (e.g., $\bfw_v^{(1)}$) to match the type of multiplicand (e.g., document embedding, $\bfx \bfu^{(1)}$).   

We note that the number of model parameters is dependent on the number of features ($m$) and documents ($n$). This implies \tgcn is not suitable for large scale applications when $(n+m)$ is very large with $n \gg m$. Furthermore, when the number of labeled training examples is small, it is difficult to learn a large number of model parameters reliably, resulting in poor generalization. Next, learning document embedding using (\ref{eqn:A}) makes \tgcn transductive. 

Motivated by the success of \textsc{PTE} and \textsc{TextGCN} in semi-supervised text classification, yet recognizing their limitations, we focus on designing a graph convolutional network-based approach that brings the best capabilities of \textsc{PTE} (i.e., efficient training and inductive) and \textsc{TextGCN} (i.e., superior performance) together.

\section{Proposed Approach}
\label{sec:proposedapproach}
We present our main idea of constructing a novel heterogeneous graph convolution network (\hgcn) variants with individual graphs used in \tgcn. Decomposition of \tgcn into different \hgcn models helps to understand the usefulness and importance of each model. \hgcn modeling approach offers flexibility in \textit{fusing} embeddings from different \hgcn models with or without \textit{layer sharing} possibilities. Furthermore, taking a cue from $\pte$, we suggest a simple method to make inference on unseen documents. Finally, we explain how \hgcn and \tgcn can be simplified by removing intermediate non-linearity and transformations, as suggested in~\cite{sgcn} and~\cite{lightgcn}.


\subsection{Heterogeneous Graph Convolutional Network (\hgcn)}
\begin{figure*}
  \centering
  \includegraphics[width=\textwidth]{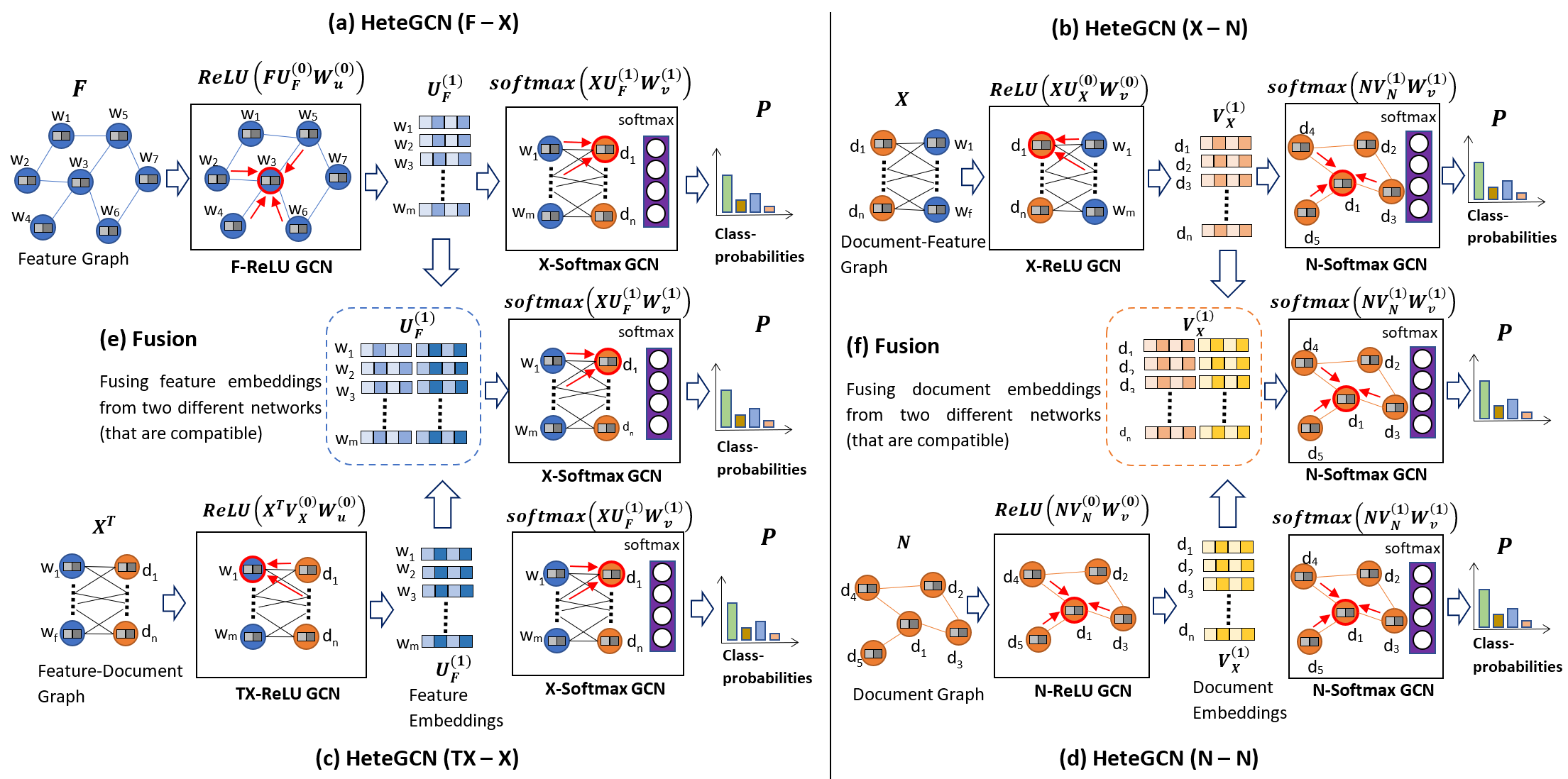}
  \caption{HeteGCN Architecture. (a),(b),(c) and (d) show networks starting from F, X, TX and N matrices, producing document embeddings, used to predict class probabilities. (e) and (f) show two possibilities of fusing feature and document embeddings coming from two different but compatible networks. $U_F^{(0)}$ and $V_N^{(0)}$ can be 1-hot representations of features and documents.}
  \Description{}
  \label{fig:hetegcn-paths}
\end{figure*}

We start with the observation that feature embeddings $\bfu^{(1)}$ are computed using both \textit{word-word} ($\bff)$ and \textit{word-document} ($\bfx$) matrices (see $\bfu^{(1)}$ in Equation~\ref{eqn:wemb}) in \tgcn. This happens mainly because of how $\bfa$ (see Equation~\ref{eqn:A}) is used in \gcn. Our proposal is to consider individual matrices $\bff$, $\bfx$ and $\bfx^T$ used in $\bfa$ separately, decompose the embedding computation operations and fuse the embeddings from different layer outputs if required. We illustrate our main idea of composing different \hgcn models in Figure ~\ref{fig:hetegcn-paths} using the following set of \gcn layers: 
\begin{itemize}
    \item \textsc{$\bff$ - ReLU \gcn}: This layer has feature embeddings as both input and output. For example, with $\bff$ in the first layer and 1-hot encoding for feature embeddings, we have: $\bfu_F^{(1)} = g(\bff \bfi_m \bfw_u^{(0)})$ with the subscript $\bff$ indicating the graph used and $g(\cdot) = \textsc{ReLU}(\cdot)$.
    \item \textsc{$\bftx$ - ReLU \gcn}: This layer takes document embeddings as input and produces feature embeddings as output. We use the prefix $\mathbf{T}$ to denote the transpose operation in $\bftx$. An example is: $\bfu_X^{(2)} = g(\bfx^T \bfv^{(1)} \bfw_u^{(1)}).$
    \item \textsc{$\bfx$ - ReLU \gcn}: This layer takes feature embeddings as input and produces document embeddings as output. In this case, we may have the second layer as: $\bfv_X^{(2)} = g(\bfx \bfu^{(1)} \bfw_v^{(1)})$.
    \item \textsc{$\bfn$ - ReLU \gcn}: This layer has document embeddings as both input and output. With $\bfn$ in the first layer and 1-hot encoding for document embeddings, we have: $\bfv_N^{(1)} = g(\bfn \bfi_n \bfw_v^{(0)})$. 
    \item \textsc{$\bfx$ - Softmax \gcn}: This layer takes feature embeddings as input and produces probability distribution $\bfp$ of documents over classes as output: $\bfp_X = \textsc{Softmax}(\bfx \bfu^{(1)} \bfw_v^{(1)})$.  
    \item \textsc{$\bfn$ - Softmax \gcn}: This layer takes document embeddings as input and produces the output:  $\bfp_N = \textsc{Softmax}(\bfn \bfv^{(1)} \bfw_v^{(1)})$.
\end{itemize}
The basic $\textsc{Softmax}$ layer produces $\bfp_I$ as output. Note that neither \pte nor \tgcn makes use of $\bfn$, while we allow the possibility of using $\bfn$ or any other graph (e.g., the \textit{word-label} relational graph used in \pte) when available. We define a homogeneous layer as a layer that consumes and produces embeddings of the same entity type (e.g., features or documents). Similarly, a heterogeneous layer consumes embeddings of one entity type (e.g., feature) and produces embeddings of another entity type (e.g., document). Thus, \textsc{$\bff$ - ReLU \gcn} is a homogeneous layer and \textsc{$\bfx$ - ReLU \gcn} is a heterogeneous layer. 

\subsection{\hgcn Models, Complexity and Implications}
Given the basic homogeneous or heterogeneous layers, the idea is to compose different \hgcn models using \textit{compatible} layers. We illustrate four \hgcn models using Figure~\ref{fig:hetegcn-paths}. In the top row (Figure~\ref{fig:hetegcn-paths}(a)), we have a \hgcn($\bff-\bfx$) model where $\bff$ and $\bfx$ are the graphs used in the first and second layer. The feature embeddings (1-hot representation at the input) and \textsc{Softmax} classifier model weights are learned using Equation~\ref{eqn:training}. Unlike \tgcn, we only learn feature embeddings. Therefore, the model complexity is $O(md + kd)$ with $d$ and $k$ denoting the feature embedding dimension and number of classes. Similarly, we have a \hgcn($\bftx-\bfx$) model (Figure~\ref{fig:hetegcn-paths}(c)) where $\bftx$ and $\bfx$ are the graphs used in consecutive layers and the document embeddings are learned and the model complexity is $O(nd + kd)$. \hgcn is heterogeneous in the sense of using different graphs across layers. This is different from traditional \gcn and \tgcn. However, two consecutive layers have to be compatible in terms of output-input relation. For example, a \textsc{$\bff$ - ReLU \gcn} layer cannot be followed by a \textsc{$\bftx$ - ReLU \gcn}. This is because the first layer produces feature embeddings as outputs and the second layer consumes document embeddings as inputs.

We observe that the model size plays an important role from two perspectives. (1) It affects the training time and tuning with a large number of hyper-parameter configurations becomes difficult. (2) When the number of labeled examples is small, it is preferable to learn models with lesser number of parameters, so that good generalization performance is achievable. We note that \tgcn learns both feature embeddings and document embeddings. Therefore, its model complexity is $O((n+m)d)$ where $d$ is the embedding dimension, assuming the same dimension for features and documents. Therefore, we may expect \hgcn models explained above to take lesser time to train and generalize well for small labeled training set scenario.

Besides model complexity, the generalization performance of each \hgcn network is dependent on the information available. For example, \hgcn($\bff - \bfx$) has rich side information (\textit{knowledge}) available through $\bff$ and may help to get significantly improved performance over \hgcn($\bfx - \bftx - \bfx$) when the number of labeled examples are small. The main reason why we may expect such performance difference is that the quality of learned feature embeddings in the former model is highly likely to be better due to feature co-occurrence knowledge. On the other hand, this knowledge has to be implicitly learned by the first two layers to derive feature embeddings in the latter mode; with a limited number of labeled examples. Similarly, \hgcn($\bff - \bfx$) may perform better than \hgcn($\bfn - \bfn$) when the quality of $\bff$ is better than $\bfn$.

It is important to note that it may not be possible to experiment with models such as \hgcn($\bff-\bfx$) due to several practical reasons. While this model may prove to be very effective, it may not always be feasible because it assumes that $\bff$ is accessible or can be explicitly computed and stored. However, $\bff$ may not be accessible or cannot be pre-computed and stored when $m$ is very large. Similarly, it may not be feasible to pre-compute $\bfn$ and store when $n$ is very large. In such situations, we are constrained to use only models such as \hgcn($\bfx - \bftx - \bfx$) or \hgcn($\bftx - \bfx$) with only available $\bfx$. 

Finally, we also note that our proposed approach is flexible. For example, we can learn a composite \hgcn model by \textit{fusing} feature embedding outputs at the layers $\bftx$ and $\bff$ of two \hgcn($\bff - \bfx$) and \hgcn($\bftx - \bfx$) models and passing the fused embedding to an \textsc{X - Softmax} \gcn as shown in the middle row (Figure~\ref{fig:hetegcn-paths}(e)). This composite \hgcn model is equivalent to performing nonlinear operation on each of the terms in $\bfu^{(1)}$ (see Equation~\ref{eqn:wemb}) separately and passing to the classifier model layer. Another possibility is to compose a composite model where the output embedding of one layer (e.g., \textsc{X - ReLU GCN}) is passed to two subsequent layers (e.g., \textsc{TX - ReLU GCN} and \textsc{N - ReLU GCN}) layers enabling \textit{shared learning} of the first layer.

\subsection{Inductive Inference} 
\label{sec:inductiveinference}
We observe that \pte is inductive because the feature embeddings are pre-computed and used during inference on unseen documents using a \textsc{Softmax} \gcn layer. Building upon this, our simple idea is to store the feature embeddings available at the appropriate layer outputs and make inference like \pte. For example, we store feature embeddings at the $\bff$ layer for \hgcn($\bff - \bfx$). Similarly, we store embeddings at the $\bftx$ and $\bfx - \bftx$ layers for \hgcn($\bftx-\bfx$) and \hgcn($\bfx-\bftx-\bfx$) models respectively. It is easy to see that the same idea can be used with \tgcn by storing $\bfv^{(1)}$. Note that any out-of-vocabulary features found in the test data will be ignored as the corresponding embeddings are not available. Experimental results in Section 4 show that this simple idea works reasonably well in practice. However, designing a good inductive method for small labeled set scenario is still a challenge when feature embeddings in the unseen data cannot be learned or are not explicitly available from other data sources (e.g., \textsc{GloVe}~\cite{glove} or \textsc{Word2Vec}~\cite{word2vec}). 

We note that the above method does not make \tgcn inductive in a strict sense  because the way document embedding is inferred during testing and training is different, as no 1-hot document representations are available for test documents -- the same holds for \hgcn($\bftx - \bfx$). Conversely, the scenario is quite different for \hgcn models with the first layer learning only the feature embeddings (e.g., $\bff-\bfx$ \:and $\bfx-\bftx- \bfx$). For these paths, it is possible to update $\bff$ and $\bfx$ as learned model weights are used for inference on new documents. $\bff$ and $\bfx$ may require updates because \textsc{IDF} factors or feature co-occurrence based $\bff$ may change. However, the quality of inference with updated matrices will be dependent on the degree of changes and the sensitivity of the learned \hgcn model with respect to the changes. More analysis and investigations are needed and are left as future work.

\subsection{Simplifying \hgcn and C-Light \gcn (\textsc{C}-\lgcn) for classification} 
\label{sec:lgcn}
There has been some effort~\cite{sgcn, lightgcn} to simplify \gcn model by understanding the usefulness of feature and nonlinear transformations.~\cite{sgcn} presented some evidence that \textit{feature smoothing} (i.e., $\bfa \bfx$) is the most important function in \gcn and performance degradation due to removal of nonlinear activation is not much. These experiments were conducted using several benchmark datasets for a classification task. The simplified \gcn classifier model turns out to be linear with a smoothed feature matrix ($\bfa^L \bfx$) fed as input where $L$ denotes the number of \gcn layers. The same idea can be used to simplify all \hgcn models (i.e., removing non-linearity). 

Using a similar idea, \lgcn~\cite{lightgcn} proposed to learn embeddings for \textit{users} and \textit{items} (starting with 1-hot representation) for recommendation problems and demonstrated that dropping both nonlinear and feature transformation operations does not degrade performance much. The resultant model is linear and a weighted linear combiner model was suggested to fuse embeddings of different layers (i.e., $\bfv =  \sum_{l=0}^{L-1} \alpha_l \bfv^{(l)}$) where $\bfv^{(l)} = \bfa^{(l)} \bfu^{(0)}$ and the weights are tuned by treating them as hyper-parameters. Like \tgcn, \lgcn uses $\bfa$ (see Equation~\ref{eqn:A}) with $\bfx$ representing the \textit{user-item} interaction matrix but without $\bff$. One simple idea is to extend \lgcn for the text classification problem by using only off-diagonal block matrices (i.e., $\bfx$) in $\bfa$ (ref. Equation~\ref{eqn:A}). It is straight-forward to derive expression for $\bfv$. It turns out that each document embedding is $(n + m)$ dimensional with contributions coming from both \textit{document-document} and \textit{document-feature} relations. The \textsc{C}-\lgcn classification model is interesting and useful because it can be interpreted as a simplified \tgcn model and compared against simplified \hgcn models.                       
\section{Experiments}
\label{sec:experiments}
We conducted several experiments to bring out different aspects of the proposed approach in comparison with the baselines and state-of-the-art graph embedding methods and answer several research questions related to network architecture. 
\subsection{Experimental Setup}
\subsubsection{Datasets}
We consider five benchmark datasets for our experiments. \textbf{20NG} consists of long text documents, categorized into 20 newsgroups. \textbf{MR} consists of movie reviews classified into positive and negative sentiments. \textbf{R8} and \textbf{R52} consist of documents that appear on Reuters-21578 newswire grouped into 8 and 52 categories, respectively. \textbf{Ohsumed} consists of medical abstracts, which are categorized into 23 cardiovascular diseases.

\subsubsection{Dataset Preparation}
For consistency, we prepare the datasets in an identical setup as described in \tgcn\cite{textgcn}. We leveraged the code\footnote{\url{https://github.com/yao8839836/text_gcn}} provided by the authors to prepare the dataset. Each of the raw text is cleaned and tokenized. Stop words and low-frequency words (less than 5 occurrences) are removed in 20NG, R8, R52 and Ohsumed except for MR, as they are short text. The statistics of the pre-processed datasets are detailed in Table~\ref{tab:dataset_stats}.

\noindent\textbf{Large Labeled Data.} We use the standard train/test split. 10\% of the train documents were randomly sampled to form the val set.

\noindent\textbf{Small Labeled Data.} We do a stratified sampling (1\%, 5\%, 10\% and 20\%) of the above training documents to form small labeled sets. Additionally, we enforce that smaller labeled training documents are included in the higher labeled training set for consistency. This is repeated 5 times to create 5 splits for each label percent. We use the val/test split as above.

\begin{table*}
\centering
\resizebox{0.8\textwidth}{!}{%
\begin{tabular}{|c|c|c|c|c|c|c|}
\hline
\textbf{Dataset} & \textbf{Words} & \textbf{Docs} & \textbf{Train Docs} & \textbf{Test Docs} & \textbf{Classes} & \textbf{Avg. Length} \\ \hline
\textbf{20NG}    & 42,757 & 18,846 & 11,314 & 7,532 & 20 & 221.26 \\ \hline
\textbf{MR}      & 18,764 & 10,662 & 7,108  & 3,554 & 2  & 20.39  \\ \hline
\textbf{R8}      & 7,688  & 7,674  & 5,485  & 2,189 & 8  & 65.72  \\ \hline
\textbf{R52}     & 8,892  & 9,100  & 6,532  & 2,568 & 52 & 69.82  \\ \hline
\textbf{Ohsumed} & 14,157 & 7,400  & 3,357  & 4,043 & 23 & 135.82 \\ \hline
\end{tabular}%
}
\caption{Dataset Statistics~\cite{textgcn}}
\label{tab:dataset_stats}
\vspace{-8mm}
\end{table*}

\subsubsection{Graph Construction}
\textbf{$\bff$} is a word-word Pointwise Mutual Information (PMI) graph. \textbf{$\bfx$}  is a document-word Term Frequency-Inverse Document Frequency (TF-IDF) graph. Refer \cite{textgcn} for further details. \textbf{$\bfn$} is a document-document nearest neighbor graph constructed from $\bfx$ with top 25 neighbours obtained using cosine similarity score. PMI and IDF are computed over all the documents in the transductive setting while only train documents are used to estimate in the inductive setting. Unseen words are removed from the validation and test documents.

\subsubsection{Methods of comparison} We compared the methods given below and all models were trained by minimizing the cross-entropy loss function (\ref{eqn:training}). Methods that learn embeddings are set to learn embeddings of size 200 for consistency.

\noindent\textbf{LR}: We trained Logistic Regression classifier on the \tfidf~transformed word vectors. We tuned hyper-parameter $C$ ($l_2$- regularization) over [1e-5, 1e5] in powers of 10 on the validation set.

\noindent\textbf{\pte}: \pte~\cite{pte} learns words' embeddings, generates text embedding from word embeddings and then utilizes them to train a logistic regression model. We used the code provided by the authors\footnote{\url{https://github.com/mnqu/PTE}} to learn the embeddings with following parameters: window = 5, min count = 0, and negative samples = 5. A logistic regression model was trained with these embeddings using \textsc{LibLinear}~\cite{liblinear} where the regularization parameter was swept over [1e-4, 1e4] in powers of 10. The parameter was tuned on the validation set.

\noindent\textbf{\tgcn}: We used the code provided by the authors\footnote{\url{https://github.com/yao8839836/text_gcn}} to set up the experiments. Apart from the best configuration suggested by the authors, we swept learning rate over [1e-1, 1e-3] in powers of 10, weight decay over [1e-2, 1e-4] in logarithmic steps and dropout over [0, 0.75] in steps of 0.25.

\noindent\textbf{C-\lgcn}: \lgcn~\cite{lightgcn} adapted to classification problems (Section~\ref{sec:lgcn}) is trained with the aggregation of adjacency matrices $A$, $A^2$, and $A^3$. The relevant rows of aggregated adjacency matrix are treated as document features, and a Logistic classifier is trained from them. The hyper-parameters $\alpha_0$, $\alpha_1$, $\alpha_2$, $\alpha_3$ of LightGCN (for aggregating adjacency matrices) and the regularization hyper-parameter $C$ of Logistic classifier are tuned using the validation set. 

\noindent\textbf{\gcn}: We run \gcn~\cite{gcn} on our datasets using $\bfn$ as the adjacency graph and $\bfx$ as input features. \gcn is equivalent to \hgcn ($\bfn - \bfn$) with $\bfx$ as input. We used the code provided by the authors\footnote{\url{https://github.com/tkipf/gcn}} to run our experiments. We swept over hyper-parameters ranges as suggested in~\cite{pitfalls} (except for embedding dimension). Additionally, graph normalization was also treated as a hyper-parameter with raw graph, row normalization and symmetric normalization as options.

\noindent\textbf{\hgcn Models}: We consider $\bff - \bfx$, $\bfx - \bftx - \bfx$, $\bftx - \bfx$ and $\bfx - \bfn$ sequences of layers in our experiments. All \hgcn models were trained for a maximum of 300 epochs using Adam optimizer and training stops when there is no increase in validation accuracy for consecutive 30 epochs. Learning rate was decayed by 0.99 after every 50 epochs. Learning rate was swept over [1e-2, 1e-4], weight decay over [0, 1e2], embedding regularization over [0, 1e2] in logarithmic steps and dropout over [0, 0.75] in steps of 0.25. Graph normalization was treated as a hyper-parameter as done for \gcn.  

\subsubsection{Evaluation Metrics}

We evaluated the performance of all classifier models using \textsc{Micro-F1} and \textsc{Macro-F1} scores~\cite{iir}. We use model accuracy (\textsc{Micro-F1}) evaluated on a held-out validation set to select the best model from various hyper-parameter configurations. 

\subsection{Large Labeled Data Scenario} 

We present results obtained from our experiments on five benchmark datasets in Table \ref{tab:largedata}. We observe that thorough tuning of hyper-parameters gives us better performance for \textsc{LR}, \pte and \tgcn models than compared to performance reported in~\cite{textgcn}. For nonlinear models (i.e., \gcn, \tgcn and \hgcn), we repeated experiments for $5$ different seeds and report average performance. In~\cite{textgcn}, \tgcn gave competitive or better performance compared to many models. For this reason, we only compare the performance of our models with \tgcn. The proposed \hgcn($\bff - \bfx$) model achieves similar or slightly better performance compared to \tgcn on \textit{all} datasets, suggesting more complex \tgcn is unnecessary. The \hgcn($\bftx - \bfx$) model gives only slightly inferior performance on four datasets (\textsc{20NG, MR, R8} and \textsc{R52}) compared to \tgcn. The main reason is that models other than \hgcn($\bff - \bfx$) do not have direct access to the \textit{word-word} relational information and this information is learned only through labeled training examples. \hgcn($\bfx - \bfn$) and $\gcn$ models give similar performance and their inferior performance is due to the quality of $\bfn$. We found the \textit{document - document} matrix to be noisy in the sense that more documents belonging to different classes are connected. Finally, \lgcn performs reasonably well on \textsc{20NG, MR and R8} but is inferior to \tgcn and we believe this is due to over-smoothing with higher powers of $\bfa$ and a lack of nonlinear transformation. We found that \lgcn is sensitive to hyper-parameter tuning. We also conducted experiments with simplified \hgcn models (i.e., removing non-linearity) and observed performance closer to that of \hgcn models.                  

\begin{table}
\small{
\begin{tabular}{|c|c|l|l|}
    \hline 
    \textsc{Dataset} & \textsc{Method} & \textsc{Micro F1} & \textsc{Macro F1} \\
    \hline
    20NG & \begin{tabular}{@{}c@{}}LR \\ \pte \\ \gcn \\ \lgcn \\ \tgcn \\ \hgcn($\bff - \bfx$) \\ \hgcn($\bfx - \bftx - \bfx$) \\ \hgcn($\bftx - \bfx$) \\ \hgcn($\bftx - \bfn$)\end{tabular} & \begin{tabular}{@{}c@{}}84.76 \\ 84.37 \\ 76.32 (0.15) \\ 82.08 \\ 85.86 (0.13) \\ \textbf{87.15} (0.15) \\ 84.39 (0.15) \\ 84.18 (0.07) \\ 76.22 (0.14)\end{tabular} & \begin{tabular}{@{}c@{}}84.09 \\ 83.70 \\ 76.07 (0.14) \\ 81.27 \\ 85.30 (0.12) \\ \textbf{86.59} (0.16) \\ 83.86 (0.13) \\ 83.71 (0.06) \\ 75.99 (0.13)\end{tabular} \\
    \hline
    MR & \begin{tabular}{@{}c@{}}LR \\ \pte \\ \gcn \\ \lgcn \\ \tgcn \\ \hgcn($\bff - \bfx$) \\ \hgcn($\bfx - \bftx - \bfx$) \\ \hgcn($\bftx - \bfx$) \\ \hgcn($\bftx - \bfn$)\end{tabular} & \begin{tabular}{@{}c@{}}77.01 \\ 71.07 \\ 74.13 (1.29) \\ 75.55  \\ \textbf{77.03} (0.10) \\ 76.71 (0.33) \\ 75.48 (0.30) \\ 75.21 (0.37) \\ 74.23 (0.82)\end{tabular} & \begin{tabular}{@{}c@{}}77.01 \\ 71.06 \\ 74.12 (1.30) \\ 75.55 \\ \textbf{77.02} (0.10) \\ 76.71 (0.33) \\ 75.46 (0.32) \\ 75.20 (0.37) \\ 74.20 (0.84)\end{tabular} \\
    \hline
    R8 & \begin{tabular}{@{}c@{}}LR \\ \pte \\ \gcn \\ \lgcn \\ \tgcn \\ \hgcn($\bff - \bfx$) \\ \hgcn($\bfx - \bftx - \bfx$) \\ \hgcn($\bftx - \bfx$) \\ \hgcn($\bftx - \bfn$)\end{tabular} & \begin{tabular}{@{}c@{}} \textbf{97.26} \\ 94.66 \\ 94.10 (0.57) \\ 96.21  \\ 96.65 (0.21) \\ 97.24 (0.51) \\ 97.09 (0.15) \\ 96.98 (0.21) \\ 94.39 (0.18)\end{tabular} & \begin{tabular}{@{}c@{}} \textbf{93.32} \\ 86.91 \\ 85.29 (0.83) \\ 88.95 \\ 87.39 (1.71) \\ 92.95 (2.01) \\ 90.99 (1.08) \\ 91.84 (1.46) \\ 86.14 (0.28)\end{tabular} \\
    \hline
    R52 & \begin{tabular}{@{}c@{}}LR \\ \pte \\ \gcn \\ \lgcn \\ \tgcn \\ \hgcn($\bff - \bfx$) \\ \hgcn($\bfx - \bftx - \bfx$) \\ \hgcn($\bftx - \bfx$) \\ \hgcn($\bftx - \bfn$)\end{tabular} & \begin{tabular}{@{}c@{}}93.38 \\ 90.65 \\ 90.41 (0.54) \\ 92.37  \\ 93.80 (0.09) \\ \textbf{94.35} (0.25) \\ 92.05 (0.41) \\ 92.87 (0.60) \\ 91.93 (0.11)\end{tabular} & \begin{tabular}{@{}c@{}}68.20 \\ 59.21 \\ 61.77 (2.07) \\ \textbf{69.29} \\ 68.62 (0.84) \\ 68.42 (1.76) \\ 52.75 (2.41) \\ 57.69 (5.67) \\ 64.61 (2.05)\end{tabular} \\
    \hline
    Ohsumed & \begin{tabular}{@{}c@{}}LR \\ \pte \\ \gcn \\ \lgcn \\ \tgcn \\ \hgcn($\bff - \bfx$) \\ \hgcn($\bfx - \bftx - \bfx$) \\ \hgcn($\bftx - \bfx$) \\ \hgcn($\bftx - \bfn$)\end{tabular} & \begin{tabular}{@{}c@{}}65.87 \\ 60.00 \\ 62.23 (0.32) \\ 65.84  \\ \textbf{68.11} (0.19) \\ \textbf{68.11} (0.70) \\ 61.44 (1.31) \\ 66.14 (0.45) \\ 62.46 (0.20)\end{tabular} & \begin{tabular}{@{}c@{}}53.77 \\ 53.67 \\ 54.87 (0.44) \\ 59.23 \\ 60.61 (0.22) \\ \textbf{60.62} (1.54) \\ 48.23 (4.17) \\ 58.52 (2.31) \\ 54.59 (0.36)\end{tabular} \\
    \hline
\end{tabular}}
\caption{\label{tab:largedata} Test Micro and Macro F1 scores on the text classification task in the large labeled setting. Models with random initializations were run with 5 different seeds, and standard deviations are reported for them in bracket wherever applicable.}
\vspace{-15mm}
\end{table}

\subsection{Small Labeled Training Data Scenario}
\begin{figure*}%
    \centering
    \subfigure{%
        \label{fig:20ng-vary}%
        \includegraphics[height=2in]{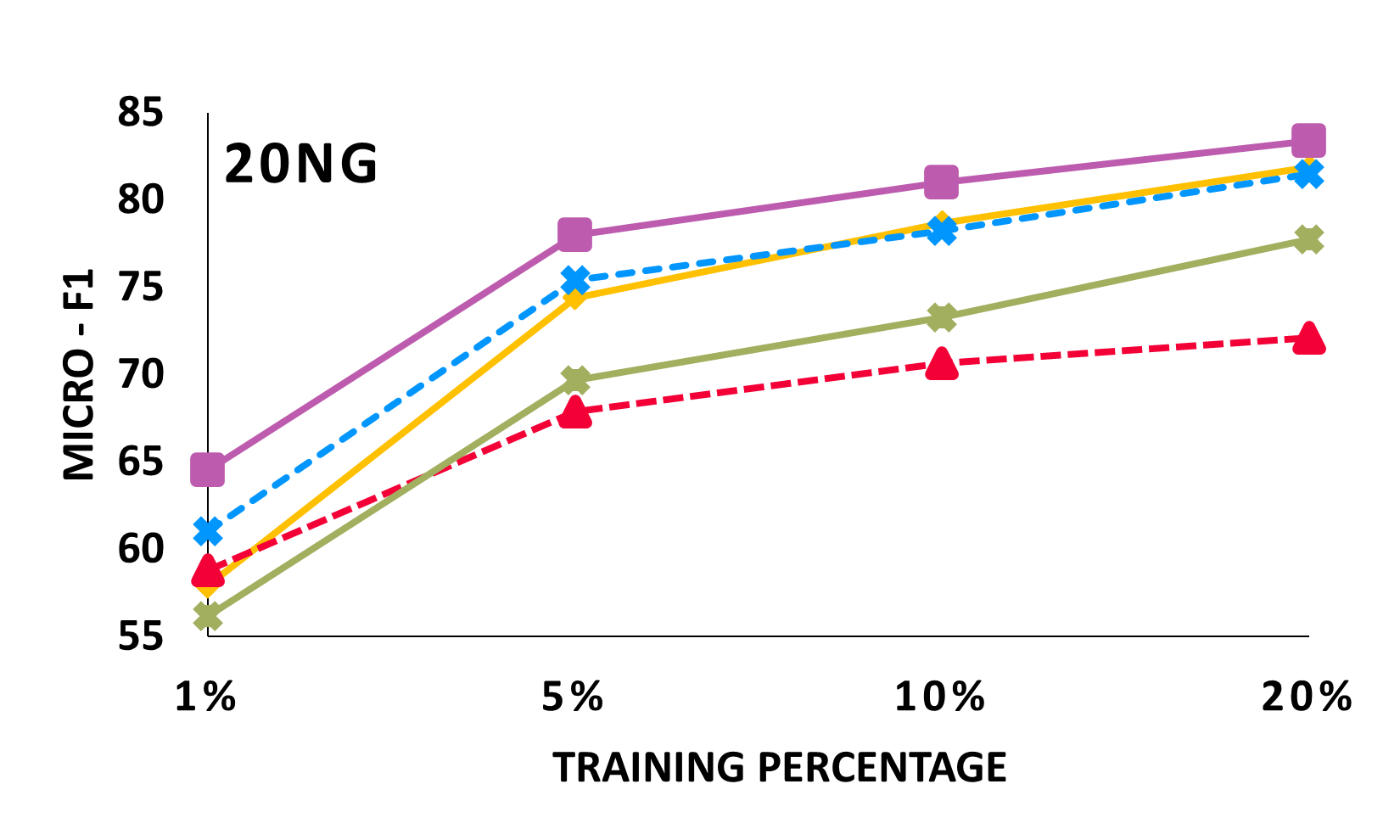}
    }%
    \hspace{8pt}%
    \subfigure{%
        \label{fig:mr-vary}%
        \includegraphics[height=2in]{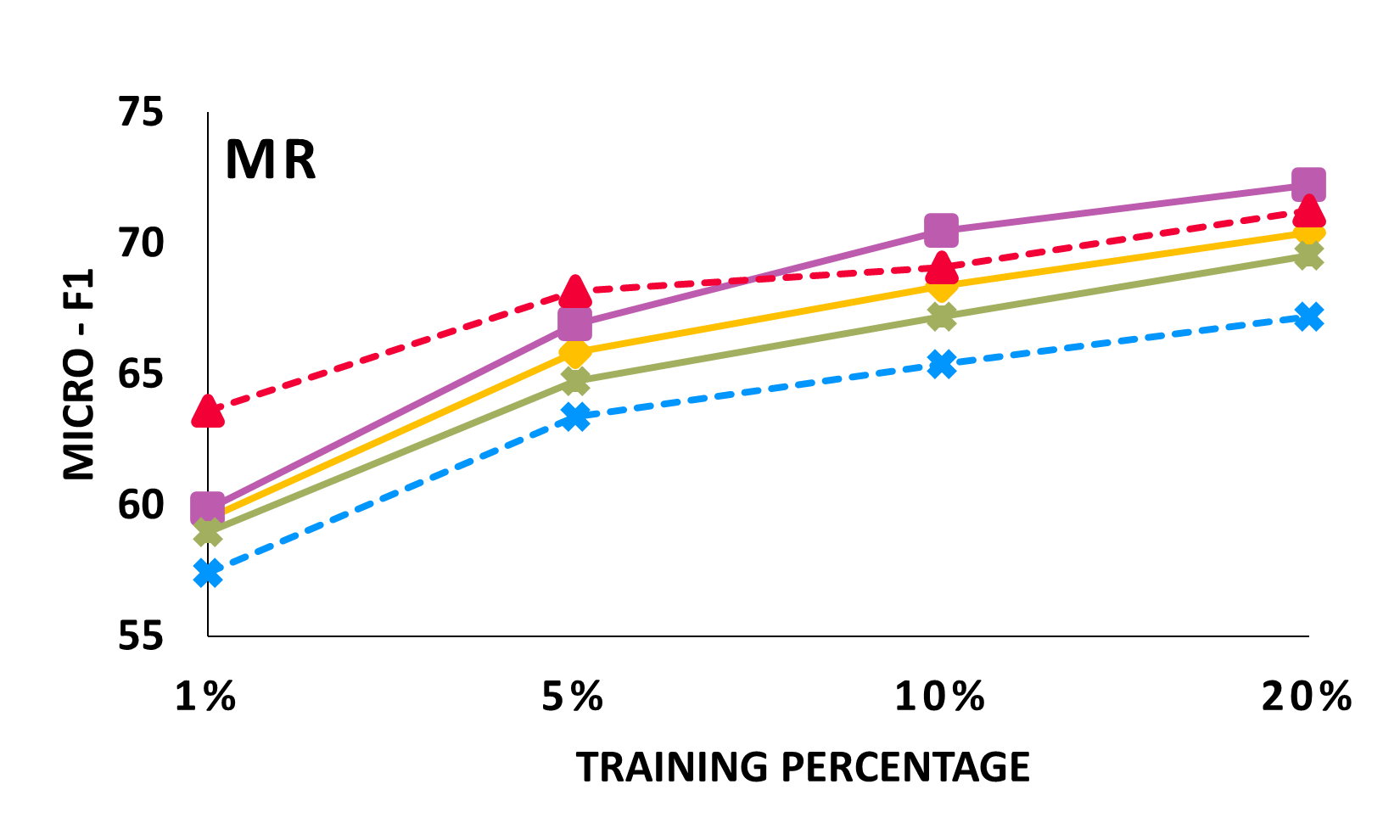}
    } \\
    \vspace{-15mm}
    \subfigure{%
        \label{fig:R8-vary}%
        \includegraphics[height=2in]{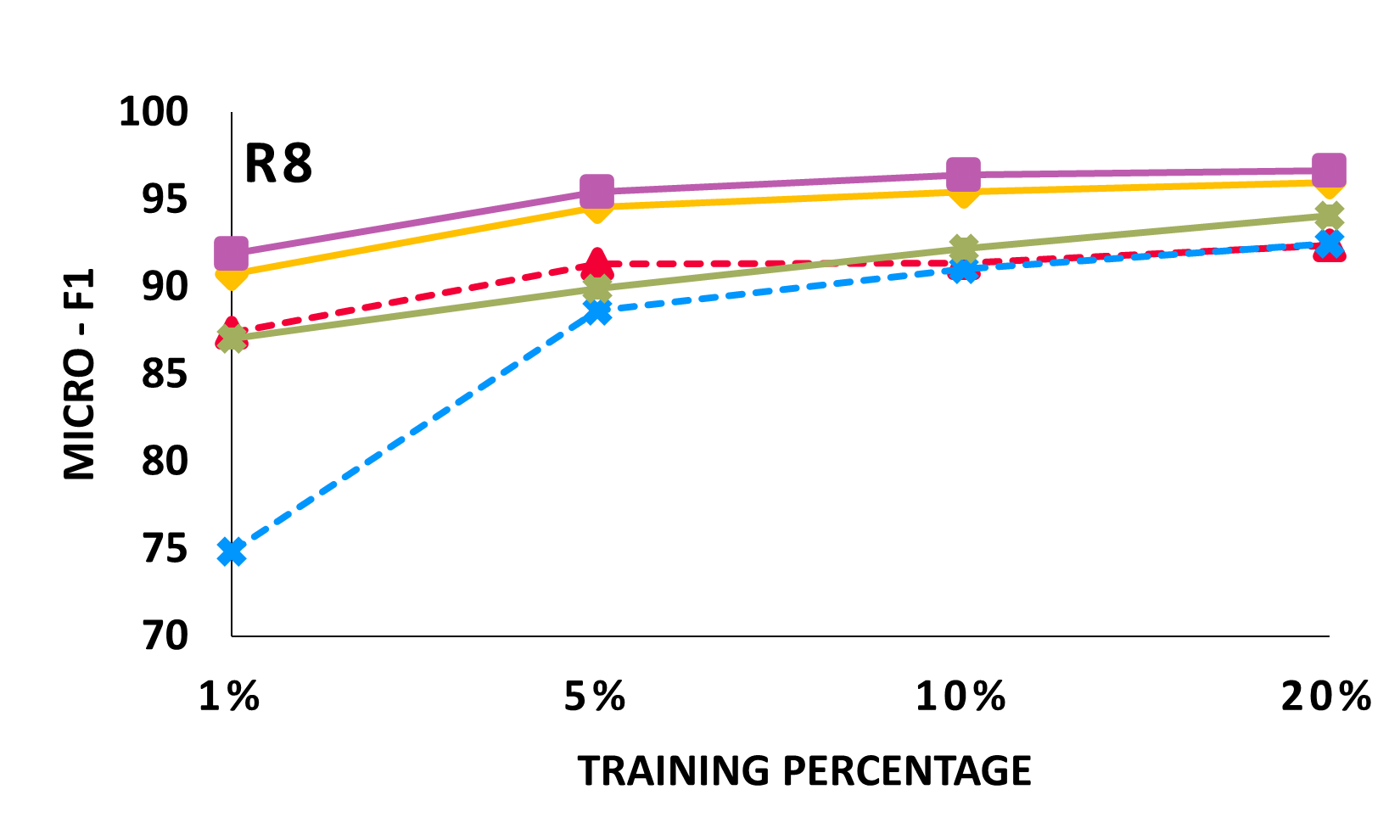}
    }%
    \hspace{8pt}%
    \subfigure{%
        \label{fig:ohsumed-vary}%
        \includegraphics[height=2in]{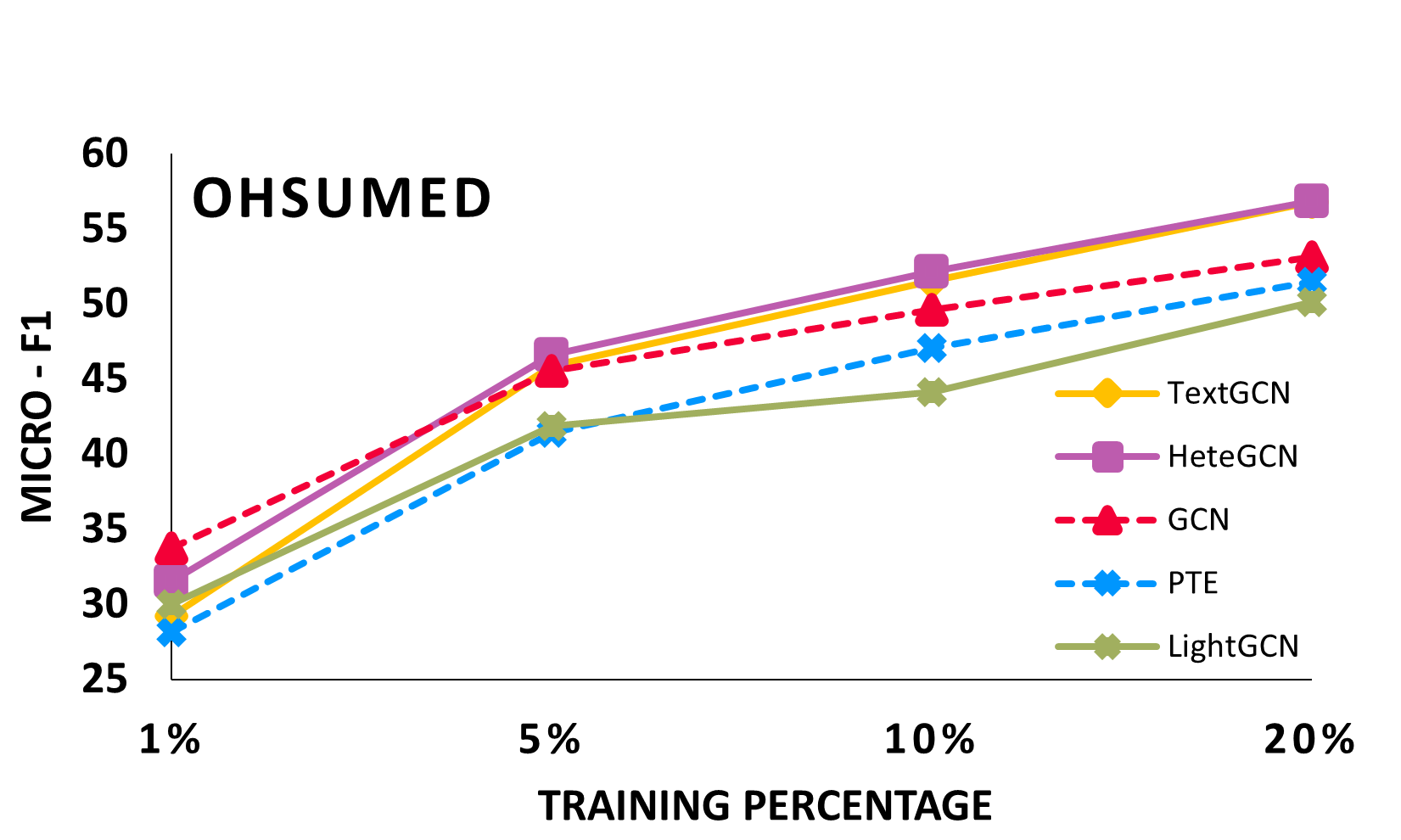}
    }%
    \vspace{-5mm}
    \caption[Varying Labeled Plots]{
        Test Micro-F1 on the text classification task plotted by varying training data sizes. The training data sizes are varied in steps of 1\%, 5\%, 10\% and 20\%.
    }%
    \label{fig:vary-train}%
\end{figure*}

We report our results from small labeled training data scenario specific experiment in Figure~\ref{fig:vary-train}.  We see that \hgcn($\bff - \bfx$) performs significantly better than \tgcn and other models in almost all datasets and varying percentage of labeled data. The performance gains are in the range of $2 - 8\%$. The superior performance of \hgcn($\bff - \bfx$) can be attributed to: (1) learning with lesser number of model parameters, (2) using $\bff$ information and (3) neighborhood aggregation and non-linear transformation advantages of \gcn. We find that the performance gap reduces as the percentage labeled data increases in several cases. \gcn and \lgcn perform better compared to \pte on \textsc{MR} and \textsc{R8}. $\pte$ performed reasonably well only in \textsc{20NG}. Note that we show only \hgcn($\bff - \bfx$) results and do not include results obtained from other \hgcn models to make Figure~\ref{fig:vary-train} clutter-free. However, the observations made on other \hgcn models in Table~\ref{tab:largedata} with respect to rest of the models (i.e., \tgcn, \pte, etc.) nearly hold even in this scenario. We also observed that the performance of \hgcn($\bfx - \bftx - \bfx$) model is close to $\tgcn$ (within $1\%$) when the percentage of labeled examples is very small (i.e., $1\%$ and $5\%$. Thus, \hgcn($\bfx - \bftx - \bfx$) model is a useful alternative to \hgcn($\bfx - \bftx$) when (1) there are memory and compute constraints related to $\bff$ and (2) we have very small labeled set. We also conducted experiments with \hgcn ($\bff-\bfx$) model with \pte embeddings fed as input. This model has lower model complexity ($O(d^2+kd)$) and gave performance lift of $(2-4\%)$ on \textsc{20NG} and \textsc{MR} datasets for $1\%$ and $5\%$ cases.    

\subsection{Inductive Experimental Study}
We conducted this experiment on large labeled data and performed inference using the inductive inference method explained in Section~\ref{sec:inductiveinference}. Results in Table~\ref{tab:indinf} show that this method is effective and useful even for \tgcn. We find that \hgcn($\bff - \bfx$) generalizes significantly better achieving ($3 - 7\%$) lifts on all datasets (except \textsc{MR}) compared to \tgcn and outperforms \pte on all datasets. We find that \hgcn($\bfx - \bftx - \bfx$) gives $1-3\%$ improvement on all datasets over \tgcn except \textsc{20NG} and is a useful alternative to \hgcn($\bff - \bfx$), as explained earlier. \pte gives surprisingly much lower performance than \textsc{LR} for e.g. in \textsc{MR}. We observed that as we increased the embedding dimension size in \pte, it tends to reach towards \textsc{LR} performance, suggesting that direct factorization of co-occurrence matrix may lead to a loss in information and that convolutional models are better from that perspective. 

\begin{table}
\small{
\begin{tabular}{|c|c|l|l|}
    \hline 
    \textsc{Dataset} & \textsc{Method} & \textsc{Micro F1} & \textsc{Macro F1} \\
    \hline
    20NG & 
    \begin{tabular}{@{}c@{}}LR   \\ \pte        \\ \tgcn        \\ \hgcn($\bff - \bfx$) \\ \hgcn($\bfx - \bftx - \bfx$) \end{tabular} & 
    \begin{tabular}{@{}c@{}}83.70\\ 81.61 (0.09)\\ 80.88 (0.54) \\ \textbf{84.59} (0.14)         \\ 79.83 (0.54)                 \end{tabular} &
    \begin{tabular}{@{}c@{}}83.01\\ 80.91 (0.08)\\ 80.47 (0.48) \\ \textbf{83.95} (0.12)         \\ 79.14 (0.57)                 \end{tabular} \\
    \hline
    MR & 
    \begin{tabular}{@{}c@{}}LR   \\ \pte        \\ \tgcn        \\ \hgcn($\bff - \bfx$) \\ \hgcn($\bfx - \bftx - \bfx$) \end{tabular} & 
    \begin{tabular}{@{}c@{}}\textbf{76.28}\\ 68.78 (0.18)\\ 74.60 (0.43) \\ 75.62 (0.26)         \\ 75.15 (0.31)                 \end{tabular} &
    \begin{tabular}{@{}c@{}}\textbf{76.28}\\ 68.76 (0.18)\\ 74.55 (0.48) \\ 75.62 (0.26)         \\ 75.15 (0.31)                 \end{tabular} \\
    \hline
    R8 & 
    \begin{tabular}{@{}c@{}}LR   \\ \pte        \\ \tgcn        \\ \hgcn($\bff - \bfx$) \\ \hgcn($\bfx - \bftx - \bfx$) \end{tabular} & 
    \begin{tabular}{@{}c@{}}93.33\\ 92.73 (0.14)\\ 94.00 (0.40) \\ \textbf{97.17} (0.33)         \\ 96.32 (0.60)                 \end{tabular} &
    \begin{tabular}{@{}c@{}}82.19\\ 82.94 (0.45)\\ 78.29 (0.59) \\ \textbf{92.33} (0.86)         \\ 89.32 (1.08)                 \end{tabular} \\
    \hline
    R52 & 
    \begin{tabular}{@{}c@{}}LR   \\ \pte        \\ \tgcn        \\ \hgcn($\bff - \bfx$) \\ \hgcn($\bfx - \bftx - \bfx$) \end{tabular} & 
    \begin{tabular}{@{}c@{}}90.65\\ 87.97 (0.12)\\ 89.39 (0.38) \\ \textbf{93.89} (0.45)         \\ 91.39 (0.42)                 \end{tabular} &
    \begin{tabular}{@{}c@{}}62.53\\ 53.10 (0.65)\\ 47.30 (2.09) \\ \textbf{66.53} (4.06)         \\ 51.31 (2.57)                 \end{tabular} \\
    \hline
    Ohsumed & 
    \begin{tabular}{@{}c@{}}LR   \\ \pte        \\ \tgcn        \\ \hgcn($\bff - \bfx$) \\ \hgcn($\bfx - \bftx - \bfx$) \end{tabular} & 
    \begin{tabular}{@{}c@{}}61.14\\ 56.93 (0.12)\\ 56.32 (1.36) \\ \textbf{63.79} (0.80)         \\ 59.12 (1.46)                 \end{tabular} &
    \begin{tabular}{@{}c@{}}\textbf{54.89}\\ 47.51 (0.45)\\ 36.74 (1.50) \\ 50.17 (2.33)         \\ 41.76 (3.01)                 \end{tabular} \\
    \hline
\end{tabular}}
\caption{\label{tab:indinf} Test Micro and Macro F1 scores on the text classification task in inductive setting. Models with random initializations were run with 5 different seeds, and standard deviations are reported for them in bracket wherever applicable.}
\vspace{-4mm}
\end{table}

\begin{table}[]
\centering
\resizebox{0.3\textwidth}{!}{%
\begin{tabular}{|c|c|c|}
\hline
\textbf{Avg. Time (s)} & \textbf{20NG} & \textbf{Ohsumed} \\ \hline
\textbf{\hgcn (F-X)}           & 4.23          & 1.09             \\ \hline
\textbf{\hgcn (TX-X)}          & 0.78          & 0.18             \\ \hline
\textbf{\hgcn (X-TX-X)}        & 1.17          & 0.27             \\ \hline
\textbf{\tgcn}       & 6.86          & 1.70             \\ \hline
\end{tabular}%
}
\caption{Average Per Epoch Time: \hgcn, \tgcn}
\vspace{-4mm}
\label{tab:timinganalysis}

\end{table}
\subsection{Timing Comparison}

We made training time comparisons on the various \hgcn~variants and \tgcn. All the analysis was done on a machine with Intel Xeon 2.60Ghz processor, 112 GB RAM, Ubuntu 16.04 OS, python 3.5 and tensorflow 1.14 (CPU). We are reporting average time taken for an epoch on 20NG and Ohsumed in Table~\ref{tab:timinganalysis}. We observe that we get $\sim1.5x$ speed up by using \hgcn($\bff - \bfx$), $\sim6x$ speed up by using \hgcn($\bfx - \bftx - \bfx$) and $\sim9x$ speed up using \hgcn($\bftx - \bfx)$. Similar speedups were observed for other datasets as well. The obtained speedups are proportional to the sparsity of the graphs involved ($\bff$ is denser than $\bfx$).

\subsection{Visualization of learned embeddings} 

\begin{figure*}%
    \centering
    \subfigure[][]{%
        \label{fig:hetegcn-tsne}%
        \includegraphics[height=1in]{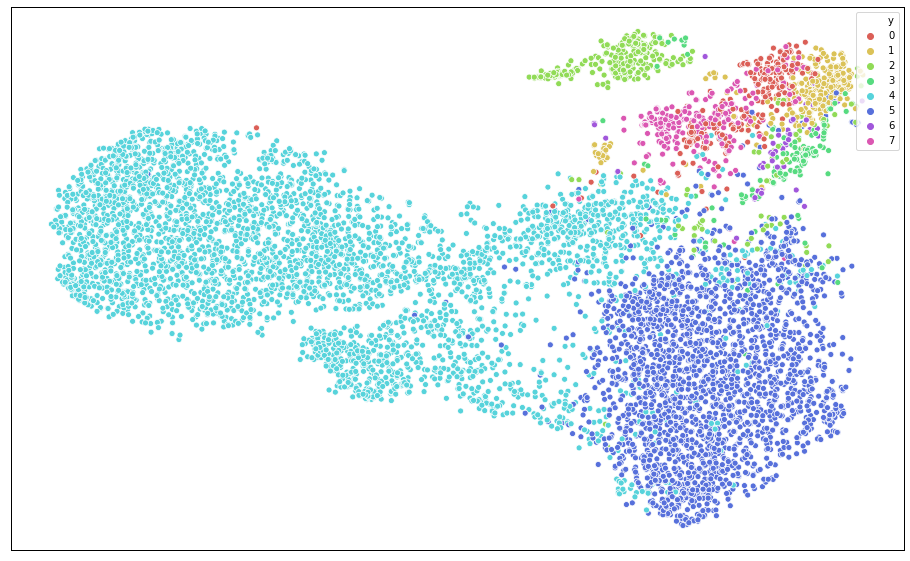}
    }%
    \hspace{8pt}%
    \subfigure[][]{%
        \label{fig:textgcn-tsne}%
        \includegraphics[height=1in]{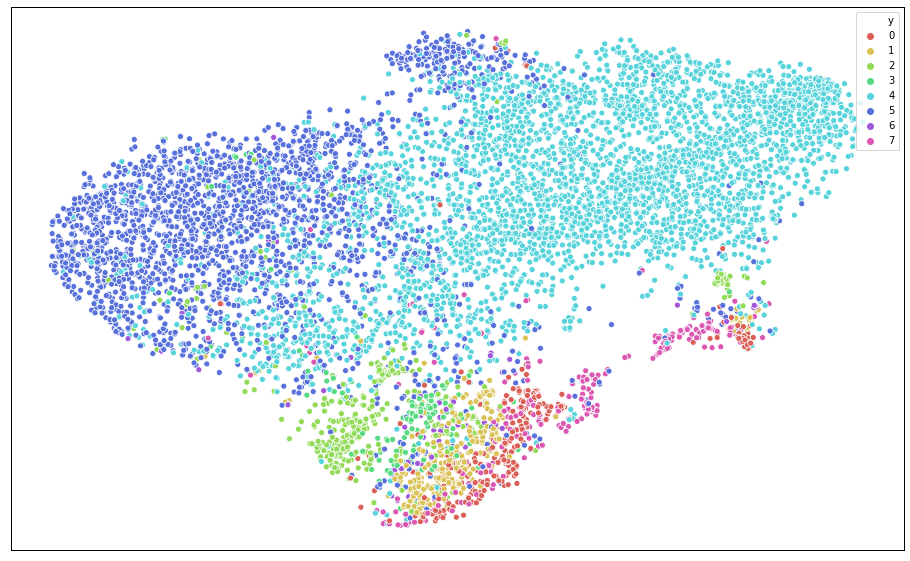}
    } 
    \subfigure[][]{%
        \label{fig:pte-tsne}%
        \includegraphics[height=1in]{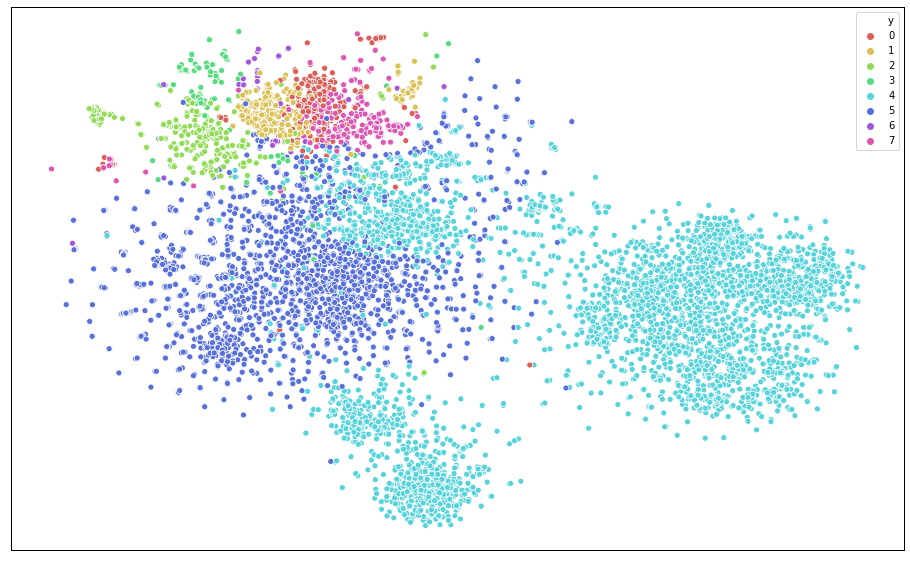}
    }%
    \hspace{8pt}%
    \subfigure[][]{%
        \label{fig:gcn-tsne}%
        \includegraphics[height=1in]{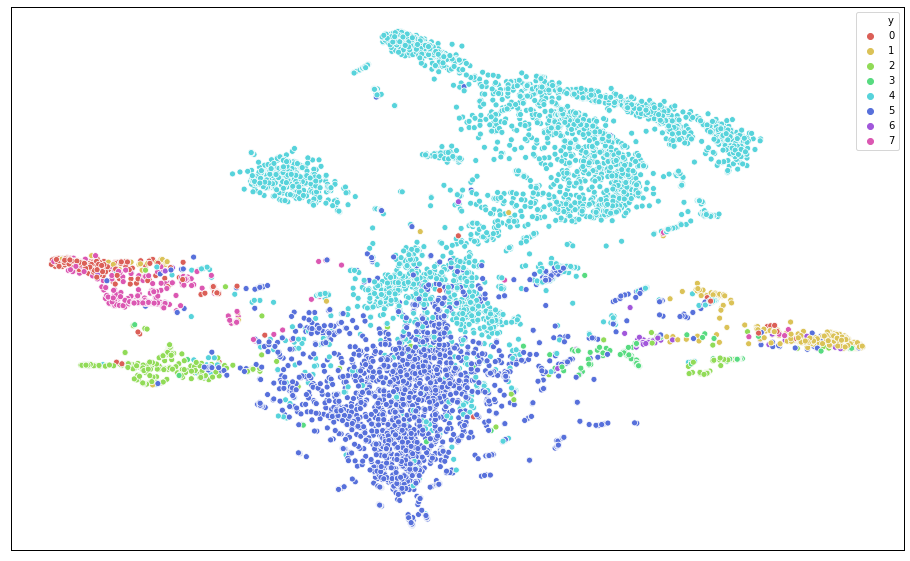}
    }%
    \vspace{-5mm}
    \caption[TSNE Plots for 20ng]{
        TSNE Plots of 20NG document embeddings obtained from the following 4 models:
        \subref{fig:hetegcn-tsne} \hgcn(F-X);
        \subref{fig:textgcn-tsne} \tgcn;
        \subref{fig:pte-tsne} \pte; and
        \subref{fig:gcn-tsne} \gcn.
    }%
    \label{fig:tsne}%
\end{figure*}

In this subsection, we discuss the effectiveness of the learned word representation. We, first, provide a tSNE~\cite{tsne} transformed document representations computed from the \hgcn trained on R8-1\% labeled dataset in Figure~\ref{fig:tsne} and compare it against those from \tgcn, \pte and \gcn.
Two things to note: (1) the majority classes (cyan and violet) are much better separated in \hgcn($\bff-\bfx$) embeddings than other models, (2) the minority classes get quite scattered in \tgcn and \gcn models, and although \pte does somewhat better, \hgcn($\bff - \bfx$) shows significantly better clustering of these points even in the 1\% labeled setting.
We also qualitatively analyse word embeddings by training a logistic regression model on aggregated training document embeddings and predicting words' labels using word embeddings. We show top-10 words with the highest probabilities for few a classes in 20NG 1\% labeled dataset in Table~\ref{tab:top10words}. We note that the top-10 words are interpretable even under low-labeled setting. 

\begin{table}[]
    \begin{tabular}{|c|c|c|c|}
        \hline
            \textbf{comp.graphics} & \textbf{sci.space} & \textbf{sci.med} & \textbf{rec.autos} \\ 
        \hline
            \begin{tabular}[c]{@{}c@{}}graphics\\ image\\ display\\ routines\\ animation\\ 3d\\ vga\\ polygon\\ processing\\ files\end{tabular} 
            & 
            \begin{tabular}[c]{@{}c@{}}space\\ moon\\ shuttle\\ orbit\\ nasa\\ mission\\ spacecraft\\ launch\\ earth\\ solar\end{tabular} 
            & 
            \begin{tabular}[c]{@{}c@{}}cancer\\ disease\\ doctor\\ patients\\ treatment\\ infection\\ drug\\ medical\\ clinical\\ medicine\end{tabular} 
            & 
            \begin{tabular}[c]{@{}c@{}}car\\ cars\\ engine\\ rear\\ ford\\ dodge\\ honda\\ gt\\ models\\ tires\end{tabular} \\ \hline
    \end{tabular}
    \caption{\label{tab:top10words} Top-10 words per class in 20NG as computed using embeddings trained on 1\% labeled data.}
    \vspace{-10mm}
\end{table}

\section{Related Work}
\label{sec:relatedwork}

Traditional text classification models~\cite{reviewcharu} and neural models~\cite{minaeereview} discussed in Section~\ref{sec:introduction} require large amount of labeled data and/or pre-trained embeddings. In practice, large labeled data is not always available. Also, raw text information might be inaccessible due to privacy concerns making it infeasible to associate any pre-trained embeddings with this data. In such cases, the various models discussed in Section~\ref{sec:introduction} cannot work effectively. 


\pte~\cite{pte} addresses these problems by learning word representations from given data by constructing a heterogeneous graph of documents, words and labels. This method can work even in a low labeled setting as long as it also has access to some unlabeled data. It can be shown that \pte factorizes a joint heterogeneous graph to learn word representations~\cite{matrixfact}. Note that utilizing unlabeled data to improve models, and using graphs to improve performance in the low labeled setting has been studied significantly~\cite{ssclds, sstcem, manifold_reg}. \pte builds on these ideas to learn better representations, while earlier models focused on improving classifier performance.

\tgcn~\cite{textgcn} combines ideas from \pte with Graph Convolutional Network (\gcn) to give better performance. \gcn~\cite{gcn} has shown excellent performance on text classification datasets. However, it assumes access to a graph structure among documents like citation networks to provide a boost in performance. Such graphs may not always be available. \tgcn, like \pte, constructs a heterogeneous graph of documents and words (excluding labels) and uses it along with \gcn. Unlike \pte, \tgcn jointly learns word representations and classifier together, thereby getting good performance. However, \tgcn has three issues: (1) it cannot scale to large datasets, (2) it force-fits a heterogeneous graph in \gcn defined for homogeneous graph, (3) it is transductive without a natural inductive formulation. Our proposed approach solves all these issues by (1) decomposing \tgcn such that the parameters are independent of the number of documents, (2) individual layers deal with one graph, either document-word or word-word, thereby ensuring consistency in the layer, giving rise to a novel heterogeneous formulation of \gcn and (3) finally the approach yields a natural inductive formulation. In this paper, we present preliminary results of the inductive formulation, and there are nuances that need further investigation.

Other recent works in text classification include~\cite{textlocalgnn, texting} and these models construct graphs at text level and use GNNs to exploit local structure in the raw text to learn text embeddings from pre-trained word embeddings. Since we assume that we do not have access to raw text information and pre-trained word embeddings, thereby, comparing against these models would not be fair. 
\section{Discussion and Future Work}
\label{sec:discussion}
In this work, we showed how different \hgcn model variants could be composed and demonstrated their effectiveness compared to complex \tgcn in different scenarios. \hgcn can be extended to work with  recommendation problems by modifying the loss function. Another application area of interest is to use \hgcn models for learning embeddings for metapaths~\cite{hine}. A metapath is defined as a sequence of entity types (e.g., \textit{user $\rightarrow$ movie $\rightarrow$ director}, \textit{user $\rightarrow$ movie $\rightarrow$ genre}) with each edge in the path specifying the relationship between entities and each metapath encodes distinct semantics. Note that embeddings for several metapaths can be learned using multiple \hgcn models with fusion and shared capabilities. Finally, deep learning models using knowledge graphs for recommendation~\cite{kgat} has been an active area of research where our approach can be used and extended with mechanisms such as \textit{attention}. We intend to explore these directions as future work.

\section{Conclusion}
\label{sec:conclusion}
We proposed a \hgcn modeling approach to construct simpler models by using \gcn layers with different graphs. \hgcn model using \textit{word-word} graph outperforms state-of-the-art models on several benchmark datasets when the number of labeled examples is small. It is quite effective in terms of model complexity, training time and performance under different labeled training data scenarios compared to \tgcn. We also suggested simpler \hgcn models that are useful when there is storage and compute constraints arising from a large number of features and documents. Finally, we also demonstrated how inductive inference is made with \hgcn and \tgcn models.  

%
\bibliographystyle{ACM-Reference-Format}
\bibliography{references}

\end{document}